# Multi-Agent System for Comprehensive Soccer Understanding

Jiayuan Rao*
SAI, Shanghai Jiao Tong University
Shanghai, China
jy_rao@sjtu.edu.cn

Zifeng Li*
Zhiyuan College & SAI, Shanghai Jiao Tong University
Shanghai, China
zifengli@sjtu.edu.cn

Haoning Wu
SAI, Shanghai Jiao Tong University
Shanghai, China
haoningwu3639@gmail.com

Ya Zhang
SAI, Shanghai Jiao Tong University
Shanghai, China
ya_zhang@sjtu.edu.cn

Yanfeng Wang
SAI, Shanghai Jiao Tong University
Shanghai, China
wangyanfeng622@sjtu.edu.cn

Weidi Xie
SAI, Shanghai Jiao Tong University
Shanghai, China
weidi@sjtu.edu.cn

## Abstract

Recent advances in soccer understanding have demonstrated rapid progress, yet existing research predominantly focuses on isolated or narrow tasks. To bridge this gap, we propose a comprehensive framework for holistic soccer understanding. Concretely, we make the following contributions in this paper: (i) we construct **SoccerWiki**, the first large-scale multimodal soccer knowledge base, integrating rich domain knowledge about players, teams, referees, and venues to enable knowledge-driven reasoning; (ii) we present **SoccerBench**, the largest and most comprehensive soccer-specific benchmark, featuring around 10K multimodal (text, image, video) multi-choice QA pairs across 13 distinct tasks; (iii) we introduce **SoccerAgent**, a novel multi-agent system that decomposes complex soccer questions via collaborative reasoning, leveraging domain expertise from SoccerWiki and achieving robust performance; (iv) extensive evaluations and comparisons with representative MLLMs on SoccerBench highlight the superiority of our agentic system.

## 1 Introduction

Sports have long been a cornerstone of human culture, captivating global audiences with their dynamic nature and emotional intensity. Among them, soccer, widely celebrated as "the beautiful game", holds a particularly prominent position, engaging billions of fans worldwide. Recent advances in artificial intelligence (AI) are transforming soccer understanding and viewing experiences by enabling automated tactical analysis [47, 53] and enriching fan engagement through automatic content generation [38, 40, 43, 44].

Generally, existing research in soccer understanding still faces two challenges: **(i) limited focus on reasoning tasks**: existing work primarily focuses on visual perception tasks, such as action spotting [7, 13] and foul recognition [21, 22], which solely rely on visual content analysis. However, reasoning tasks often require the assistance of extra context or knowledge, for example, answering *"How many goals and assists did this ball-carrying player make in the 2019-2020 season?"* would require both visual athlete identification and knowledge retrieval; **(ii) fragmented and specialist models**: most studies typically develop specialist models for isolated tasks, which can be potentially labor-intensive and challenging to scale. Heterogeneous annotation formats across distinct tasks further impede the development of generalist models and comprehensive evaluations, contrasting with modern AI research paradigms that emphasize generalization and adaptability.

In this work, we introduce the task of knowledge-based question-answering for comprehensive and standardized assessment of soccer understanding. Given the reliance on soccer domain knowledge, we first construct **SoccerWiki**, a large-scale multimodal soccer-specific knowledge base, comprising extensive information about 9,471 players, 266 teams, 202 referees, and 235 venues from the Internet. By integrating SoccerWiki and various soccer datasets [5, 7, 22, 38, 43, 44] through an automated data curation pipeline and manual verification, we establish **SoccerBench**, the largest and most comprehensive benchmark for soccer understanding to date, featuring around **10K** multi-choice QA samples across **13** soccer analysis tasks, including *background knowledge, match situation, camera status classification, jersey number recognition, jersey colors, camera status switching, replay grounding, action classification, commentary generation*, and *multi-view foul recognition.*

Despite significant advances in Multimodal Large Language Models (MLLMs) [3, 27, 33, 50], soccer understanding remains challenging due to its complexity and knowledge-intensive nature. General-purpose MLLMs, constrained by their limited soccer-specific prior knowledge, struggle to address the diverse and highly specialized questions posed in **SoccerBench**. To tackle this, we propose **SoccerAgent**, a novel multi-agent system, as illustrated in Figure1.

*These authors contribute equally to this work.

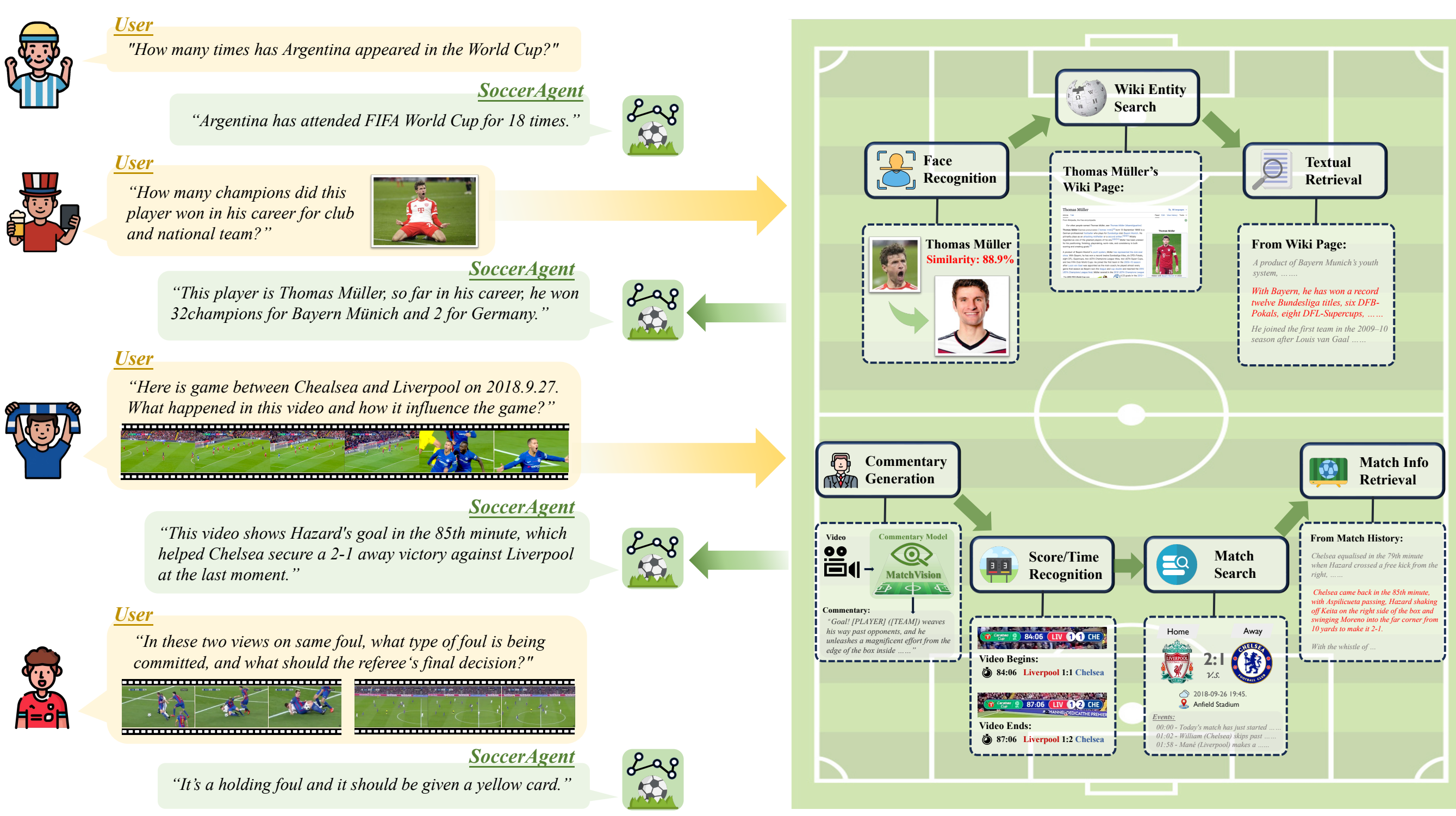

(a) Examples of SoccerBench and our Interface

(b) Examples of SoccerAgent Reasoning Chain

**Figure 1: Overview. (a) A user example of our multi-agent system, SoccerAgent, on the proposed diverse and challenging SoccerBench; (b) An example of the reasoning chain and workflow of SoccerAgent.**

Concretely, SoccerAgent leverages a powerful agent core [32] capable of invoking 18 specialized tools (with 17 of them being **open-source**). Given a soccer-related question, the agent system begins by decomposing the task into multiple sub-tasks executable by existing tools, and then invokes corresponding tools to process, capturing both fine-grained player actions and macro-level team strategies for comprehensive soccer analysis.

To summarize, we make the following contributions in this paper: (i) we present **SoccerWiki**, the first large-scale multimodal knowledge base for soccer, comprising extensive information about 9,471 players, 266 teams, 202 referees, and 235 venues, enabling knowledge-based question answering beyond simple visual perception; (ii) we construct **SoccerBench**, the largest and most comprehensive soccer-specific benchmark through an automated curation pipeline and manual verification, featuring around 10K standardized multi-choice QA pairs across 13 soccer-specific tasks; (iii) we develop **SoccerAgent**, a novel multi-agent system where specialized agent tools collaborate to integrate domain knowledge and achieve robust soccer game understanding; (iv) we conduct extensive evaluations against 11 representative MLLMs, demonstrating the challenging characteristics of SoccerBench and the superiority of SoccerAgent. We believe these will establish a foundation for future research in evolving, knowledge-driven sports analytics.

## 2 Related Works

### 2.1 Question Answering in Vision

Question Answering (QA) serves as a pivotal testbed for assessing the comprehensive understanding capabilities of Multimodal Large Language Models (MLLMs). Early datasets like VQA [2] and COCO-Caption [4], primarily focus on narrow tasks (*e.g.*, object recognition or image captioning), while recent advancements [10, 20, 26, 37, 56, 64, 67] have significantly broadened the evaluation scope. MME [9] integrates 14 perception/cognition tasks, and MMBench [36] improves robustness via ChatGPT-based answer alignment. MMMU [63] further advances by covering university-level problems across six disciplines (*e.g.*, engineering, arts), demanding expert knowledge synthesis from multimodal inputs.

### 2.2 Sports Understanding

Sports understanding [51] is an emerging field that integrates multiple data modalities across various disciplines, encompassing diverse tasks such as automated scoring [45, 60], action spotting [6, 7, 13, 16], foul recognition [21, 22], commentary generation [38, 40, 43, 44], and tactical analysis [47, 53, 61]. While prior works typically focus on developing specialized models for individual tasks, recent advances in MLLMs have enabled more holistic sports understanding evaluation [29, 55, 58, 59]. In this paper, we focus on soccer, the most popular sport worldwide, and construct the largest and most comprehensive multimodal soccer-specific benchmark to date, aiming to promote development in this field.

**Table 1: Data Statistics of SoccerBench. For each, we present its name, QA type, source materials, and curation strategies. Here, SN and SR-1988 represent the SoccerNet and Soccer-Replay-1988, respectively, while LLM denotes DeepSeek-v3 [32].**

| Index | Task | Type | #Samples | Data Source | Materials | Curation |
|---|---|---|---|---|---|---|
| Q1 | Background Knowledge Text QA | Text | 1,500 | SoccerWiki | - | LLM |
| Q2 | Match Situation QA | Text | 1,200 | SoccerWiki | - | LLM |
| Q3 | Camera Status Classification | Image | 400 | SN-v2 [7] | 400 images | Template |
| Q4 | Background Knowledge Image QA | Image | 1,000 | SoccerWiki | 2,235 images | LLM |
| Q5 | Jersey Number Recognition | Image | 200 | SN-JN [5] | 99,252 images | Template, LLM |
| Q6 | Score and Time Relevant QA | Image | 600 | SN-Caption [38], SR-1988 [43] | 633 images | Template, LLM |
| Q7 | Camera Status Switching | Video | 400 | SN-v2 [7] | 400 videos | Template |
| Q8 | Replay Grounding | Video | 400 | SN-v2 [7] | 2,105 videos | Template |
| Q9 | Action Classification | Video | 1,000 | SN-v2 [7], MatchTime [44], SR-1988 [43] | 1,000 videos | Template |
| Q10 | Commentary Generation | Video | 1,000 | SN-Caption [38], SR-1988 [43] | 1,000 videos | Template |
| Q11 | Commentary Relevant QA | Video | 800 | SN-Caption [38], SR-1988 [43] | 1,000 videos | LLM |
| Q12 | Jersey Color Relevant QA | Video | 700 | SoccerWiki, SR-1988 [43] | 700 videos | LLM |
| Q13 | Multi-view Foul Recognition | Video | 300 | SN-XFoul [21] | 435 videos | Template |

## 2.3 Multi-Agent System

Multi-Agent System (MAS) has emerged as a powerful paradigm for modeling complex interactions among autonomous entities, with applications across language model collaboration [18, 30, 62], embodied AI [19, 49, 54], and scientific problem-solving [8, 12]. Recent advances [23, 46, 52] highlight the effectiveness of MAS in multimodal reasoning tasks through role specialization and inter-agent communication. Frameworks like CAMEL [28], AutoGen [57], and ChatDev [41] have further refined agent communication protocols for complex task decomposition. This paper presents **SoccerAgent**, the first soccer-specific multi-agent system that decomposes complicated questions through collaborative reasoning and accomplishes 13 distinct soccer understanding tasks.

# 3 Dataset Construction

This section outlines the motivation and overview of our dataset in Sec. 3.1, followed by detailed descriptions of the data collection and curation processes in Sec. 3.2 and Sec. 3.3, respectively.

## 3.1 Motivation & Overview

Soccer is a dynamic and specialized domain, with its evolving nature often outpacing the static knowledge encoded within pre-trained multimodal large language models (MLLMs). To bridge this gap, we introduce **SoccerWiki**, a dynamic, large-scale knowledge base that provides up-to-date and comprehensive information on players, teams, referees, and venues. SoccerWiki spans data from the past decade of the **top five European Leagues, the UEFA Champions League, and the last three FIFA World Cups**.

While existing research in soccer AI primarily targets isolated tasks, it lacks a holistic framework for comprehensive evaluation. To address this, we present **SoccerBench**, a multimodal benchmark for soccer understanding. By integrating SoccerWiki with various existing datasets [7, 22, 38, 43, 44] through an automated curation pipeline, SoccerBench unifies 13 distinct soccer-specific analysis tasks into a standardized question-answering (QA) framework. It includes approximately 10,000 QA pairs, enabling a robust and comprehensive evaluation of soccer understanding models.

## 3.2 Data Collection

To construct a diverse multimodal soccer-specific knowledge base, **SoccerWiki**, we aggregate comprehensive soccer-related information from Wikipedia[1] and Flashscore[2], **covering 9,471 players, 266 teams, 202 referees, and 235 venues**. Each entity in the knowledge base includes the corresponding image and detailed attributes, such as career statistics, personal profiles, team histories, and honors. Additionally, we have incorporated detailed game information from 1,988 soccer matches (from six major European soccer leagues and championships) in the SoccerReplay-1988 [43] dataset, covering team lineups, key event annotations, and detailed captions. To further improve data coverage, we have manually annotated the jersey colors for both home and away teams of these matches. Notably, SoccerWiki can dynamically update the up-to-date soccer information by leveraging real-time match data from Flashscore and integrating information from the Wikipedia API.

To formulate a soccer-specific multimodal benchmark, namely, **SoccerBench**, under an unified question-answering framework, we leverage extensive data from SoccerWiki and annotations from various existing soccer datasets, including: (i) textual commentary from SoccerReplay-1988 [43] and SoccerNet-Caption [38], (ii) event labels from SoccerReplay-1988 [43], SoccerNet-v2 [7] and SoccerNet-test-align [44], (iii) foul classification labels from SoccerNet-XFoul [22], (iv) jersey number labels from SoccerNet-JN [5], (v) camera status and replay labels from SoccerNet-v2 [7]. The comprehensive integration of these diverse data sources and tasks ensures the exceptional coverage and challenging characteristics of SoccerBench.

## 3.3 Data Curation

*3.3.1 Open-ended QA Construction.* As depicted in Table 1, we categorize the extensive data collection into 13 subtasks based on their annotations, and construct open-ended QA pairs by employing predefined templates or prompting LLMs like DeepSeek-v3 [32]. For questions in a relatively uniform and fixed question pattern, such as "What type of event is happening in this video?", which can be handled by visual perceptions, we design questions via predefined templates according to the annotations from existing benchmarks. And for tasks requiring soccer-specific knowledge or factual data, *e.g.*, "How many teams did the player in the image play for during his career?", we adopt LLMs to create QA pairs with appropriate prompts, which will be detailed in the **Appendix**.

[1]www.wikipedia.org
[2]www.flashscore.com

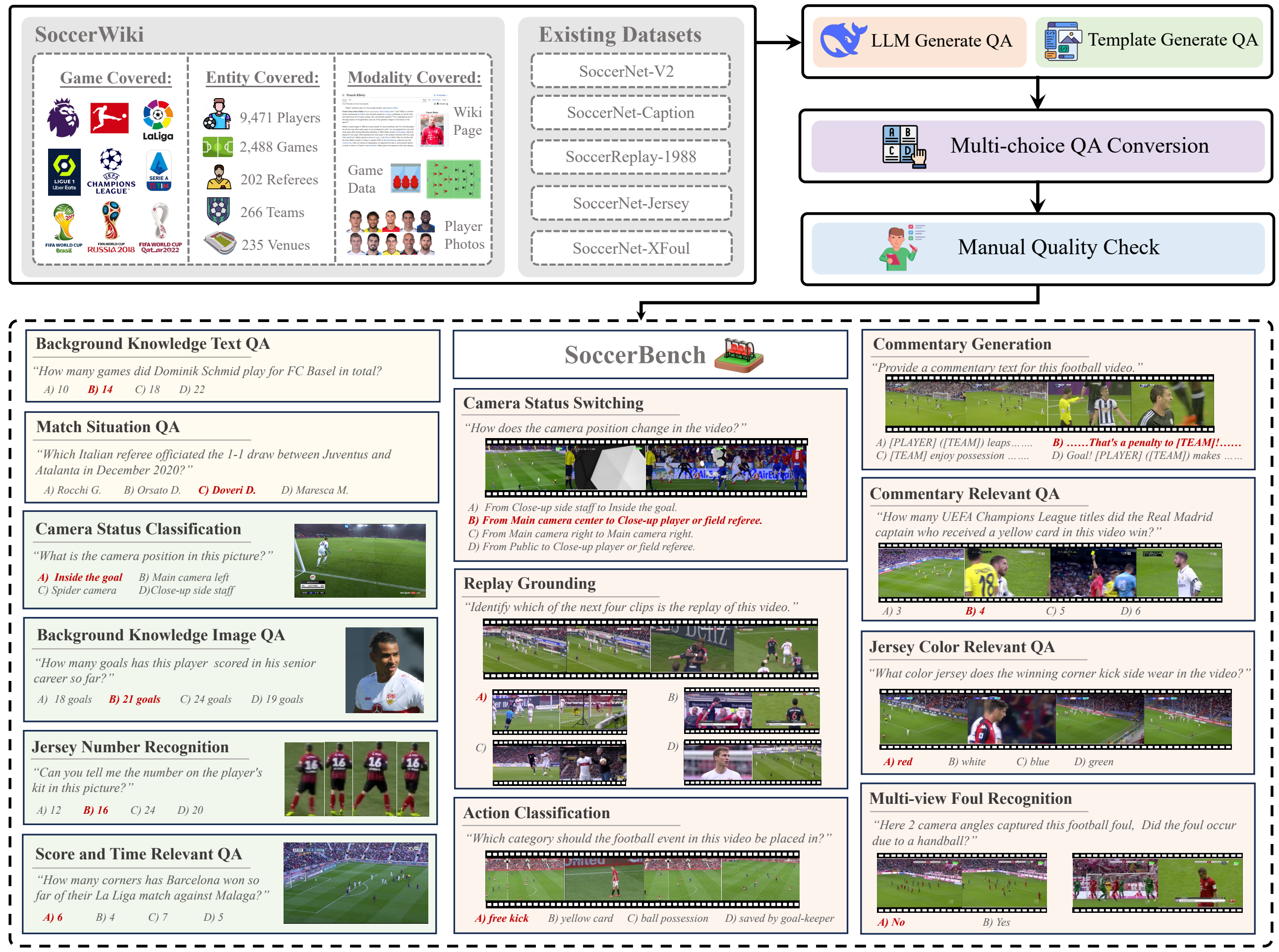

**Figure 2: SoccerBench QA Generation Pipeline. We construct multi-choice QA samples based on SoccerWiki and other existing datasets. Some representative examples for each task are presented for reference.**

*3.3.2 Conversion from Open-ended QA to Multi-choice QA.* To facilitate efficient quantitative evaluation, we convert open-ended QA pairs into multi-choice format, each with one correct answer and three carefully designed distractors. Concretely, we employ two strategies to construct plausible yet challenging distractors: (i) randomly sample labels from the same category as the correct answer (*e.g.*, action and camera); and (ii) prompt DeepSeek-v3 [32] to create distractors that may introduce confusion (*e.g.*, numbers and dates). These strategies ensure the complexity and perplexity of our challenging benchmark. Through this scalable curation pipeline, we automatically synthesize 100K QA pairs, then manually select around 10K representative samples to form **SoccerBench**. More data curation details will be provided in our **Appendix**.

*3.3.3 Discussion.* As depicted in Table 1, SoccerBench covers 13 distinct soccer-specific QA tasks with a balanced distribution. Some representative examples of each task are presented in Figure 2, showcasing the diverse formats and content across various tasks. Among them, tasks (Q1)-(Q2) are text-based QA, (Q3)-(Q6) involve image-related QA, and (Q7)-(Q13) focus on video-related QA. To the best of our knowledge, SoccerBench represents the largest and most comprehensive multimodal soccer-specific benchmark to date, covering diverse complexity levels, modalities, and task categories. We believe it will serve as a valuable resource for soccer understanding evaluation, thus advancing research in sports analysis.

## 4 Methodology

We introduce **SoccerAgent**, a multi-agent system that leverages a modular architecture for comprehensive analysis and precise responses to multimodal soccer-related questions, addressing diverse knowledge-intensive soccer understanding tasks. In this section, we begin by formulating the problem and defining the scope of soccer-related questions in Sec. 4.1; Next, we detail the specialized tools integrated into the system in Sec. 4.2; Finally, in Sec. 4.3, we elaborate on the architecture, operational logic, and interaction mechanisms that enable SoccerAgent to deliver robust results.

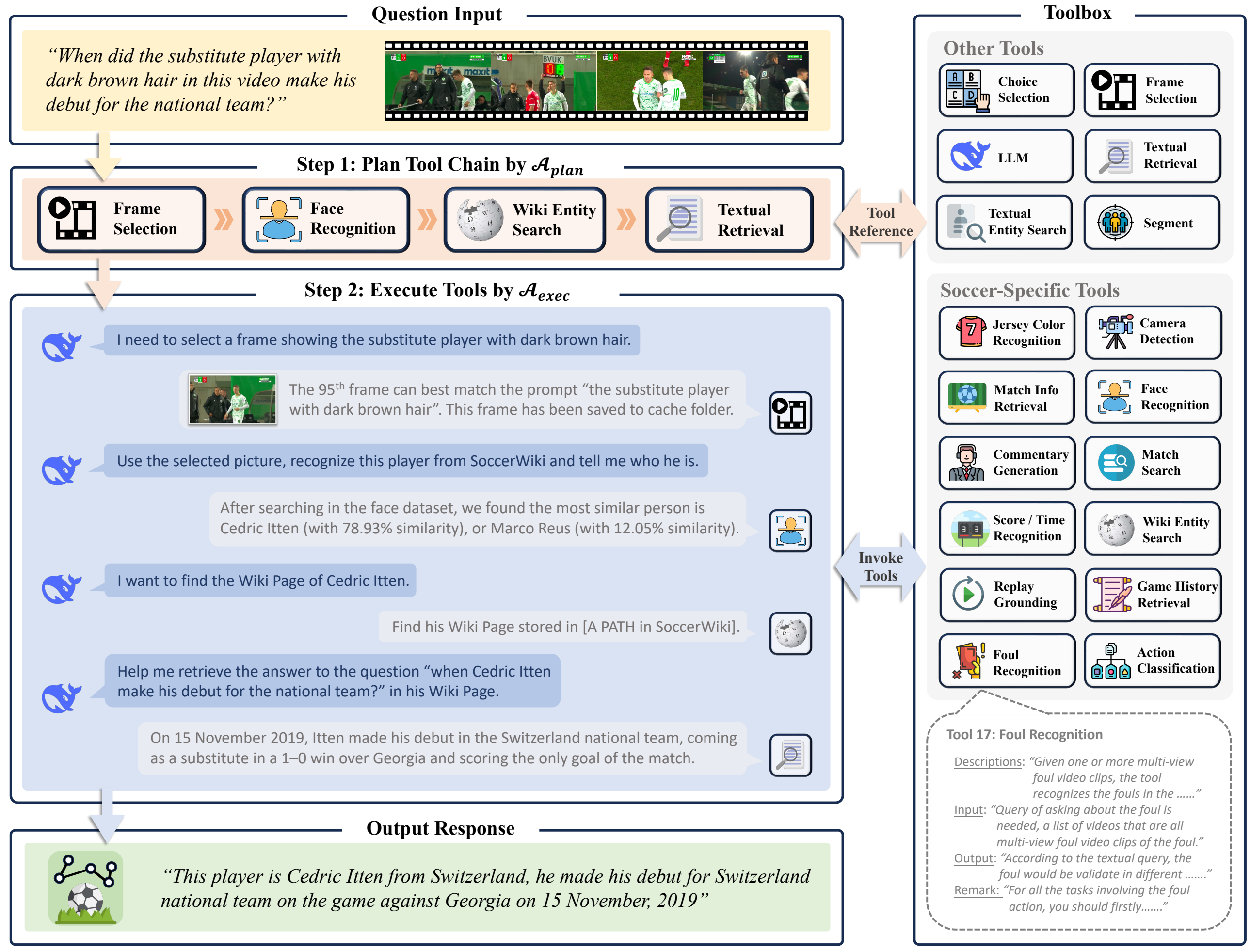

**Figure 3: SoccerAgent Architecture Overview. We design a multi-agent system to decompose and solve the given multi-modal soccer-related questions step by step with a distributed toolbox.**

## 4.1 Problem Formulation

As outlined in Sec. 3, soccer understanding tasks encompass diverse challenges that demand nuanced reasoning across visual and contextual knowledge domains. To tackle these complexities, our framework adopts a multi-agent collaborative system integrated with **existing tools and models**, ensuring adaptability, scalability, and transparency. The core workflow can be formally expressed as:

$$\mathbf{r} = \mathcal{A}(\mathbf{q}; \mathcal{T})$$

where $\mathcal{A}$ denotes our multi-agent system, **SoccerAgent**, comprising two primary modules: $\mathcal{A}_{plan}$ is responsible for planning and determining the optimal tool chain to address the input question, and $\mathcal{A}_{exec}$ executes the planned tool chain, ensuring seamless integration of outputs from individual tools. Here, we employ DeepSeek-v3 (671B) [32] as the agent core for both $\mathcal{A}_{plan}$ and $\mathcal{A}_{exec}$.

The input question ($\mathbf{q}$) represents a specific multimodal soccer-related question, while $\mathcal{T}$ refers to a dynamically configurable toolbox, expressed as $\mathcal{T} = \{\mathbf{t}_1, \mathbf{t}_2, \ldots, \mathbf{t}_n\}$, where each $\mathbf{t}_i$ represents a distinct tool. Each tool ($\mathbf{t}$) performs a specific task based on the input instruction ($\mathbf{s}$) and produces an execution output $\tau = \mathbf{t}(\mathbf{s})$. Finally, the system produces a comprehensive response ($\mathbf{r}$) by aggregating and reasoning over the outputs. Note that, the tools can be any existing APIs or models, and in this paper, **open-source frameworks are prioritized** for our toolbox wherever feasible.

## 4.2 Tools

Our toolbox integrates 18 specialized tools, each designed to handle specific functionalities across diverse modalities. These tools are rigorously defined with clear input/output specifications to guide the decision-making and operation processes of **SoccerAgent**. As depicted in Figure 3, the toolbox consists of 12 soccer-specific tools and 6 general-purpose tools, systematically categorized as follows.

*4.2.1 Off-the-shelf Soccer-specific Tools.* We adopt two tools from the pre-trained soccer understanding model proposed in UniSoccer [43]: (i) *Action Classifier*: automatically classifies actions, and (ii) *Commentary Generation*: produces anonymized textual descriptions of video content. These tools operate directly on soccer video data, providing foundational capabilities for soccer-specific analysis.

*4.2.2 Soccer-specific Retrieval Tools.* Leveraging information in SoccerWiki, we develop four retrieval tools: (i) *Match Search:* retrieves relevant match records from SoccerWiki based on textual queries; (ii) *Match History Retrieval:* extracts and summarizes event statistics from identified matches; (iii) *Match Info Retrieval:* fetches extra match details, such as referee name and line-up formations; and (iv) *Face Recognition:* identifies players by matching input images with facial photos in SoccerWiki. The first three tools are implemented by DeepSeek-v3 [32] with customized prompts , while the last tool adopts an open-source face recognition framework [11].

*4.2.3 Soccer-specific Image Understanding Tools.* Based on Qwen2.5-VL-7B [3] with carefully crafted prompts (detailed in the **Appendix**), we develop three soccer-specific image understanding tools: (i) *Camera Detection:* serves as a classifier to recognize camera position types; (ii) *Jersey Number Recognition:* first checks if the image contains jersey numbers via a pretrained model [25] and then extracts them; and (iii) *Score/Time Recognition:* captures the scoreboard and game time from broadcast images.

*4.2.4 Soccer-specific Video Understanding Tools.* Similarly, Qwen2.5-VL-7B [3] also serves as the core of video understanding tools, including: (i) *Replay Grounding:* analyzes replay clips and verifies their consistency with live broadcast footage; (ii) *Jersey Color Recognition:* recognizes the jersey color of players in the given footage and answers relevant questions; and (iii) *Foul Recognition:* functions as a multi-view video referee system, aggregating inputs from different views through a voting mechanism to determine final decisions.

*4.2.5 General-purpose Multimodal Parsing Tools.* To support generic multimodal and logical operations, we implement six general-purpose tools: (i) *Frame Selection:* adopts CLIP [42] text-to-image similarity to extract the video frame most semantically aligned with a given textual prompt, effectively converting video content into keyframes; (ii) *Segment:* leverages off-the-shelf GroundingDINO [35] to detect and localize relevant entities in images with precise bounding boxes corresponding to given text prompts. (iii) *Textual Entity Search:* extracts potential key entities (players, teams, referees, *etc.*) from input questions for subsequent processing by subsequent tool processing; (iv) *Textual Retrieval:* fetches relevant information from long-form text based on specific query prompts; (v) *Answer Selection* is specifically designed for multiple-choice scenarios, returning the most probable answer; and (vi) a general *LLM Tool:* serves as a default module for arbitrary language model operations when needed. Here, the last four tools are implemented with DeepSeek-v3 [32], enabling efficient text-based information extraction and expansion.

## 4.3 SoccerAgent

*4.3.1 Tool Chain Planning.* As depicted in Figure 3, given a specific soccer-related question ($\mathbf{q}$), the planning agent ($\mathcal{A}_{plan}$) systematically constructs an optimal tool chain ($\mathbf{C}$) through reasoning about the question's requirements and the capabilities of the tools available in the toolbox. This process can be formally expressed as:

$$\mathbf{C} = \mathcal{A}_{plan}(\mathbf{q}, \mathcal{T}) = [\mathbf{t}_i \rightarrow \mathbf{t}_j \rightarrow \cdots \rightarrow \mathbf{t}_m]$$

Here, C represents the ordered sequence of tools required to address the input question ($\mathbf{q}$), and $\mathcal{T}$ denotes the set of tools described in Sec. 4.2. The planned chain ensures that each tool contributes meaningfully to solving the task while adhering to the input/output compatibility between consecutive tools.

*4.3.2 Iterative Tool Execution.* Once the tool chain is planned, the execution agent ($\mathcal{A}_{exec}$) processes each tool in the chain iteratively, considering the original question ($\mathbf{q}$) and the accumulated execution history. At each step, the agent generates the instruction input ($\mathbf{s}_i$) based on the question ($\mathbf{q}$) and the accumulated execution history ($\mathcal{H}_i$), enabling adaptive and context-aware input selection for each tool. The execution history ($\mathcal{H}_i$) at step ($i$) is defined as:

$$\mathcal{H}_i = \{(\mathbf{t}_1, \mathbf{s}_1, \tau_1), \cdots, (\mathbf{t}_{i-1}, \mathbf{s}_{i-1}, \tau_{i-1})\}, \text{ with } \mathcal{H}_0 = \varnothing$$

The instruction input ($\mathbf{s}_i$) and the tool ($\mathbf{t}_i$) are then used to compute the output ($\tau_i$) as follows:

$$\tau_i = \mathbf{t}_i(\mathbf{s}_i), \ \mathbf{s}_i = \mathcal{A}_{exec}(\mathbf{q}, \mathcal{H}_{i-1}; \mathbf{t}_i)$$

To ensure consistency and interpretability, each instruction generated by $\mathcal{A}_{exec}$ adheres to a strictly structured format, encapsulated within `<Call></Call>` markers, with four specialized delimiters specifying key execution parameters: `<Tool></Tool>` denotes the name of the invoked tool; `<Query></Query>` contains the text input for the tool; `<Material></Material>` provides the file paths of the input visual content; `<Purpose></Purpose>` articulates the rationale and objective for executing the tool at this step.

Upon reaching the final execution step, $\mathcal{A}_{exec}$ makes the last execution in `<EndCall></EndCall>` markers, the output generated by this terminal step, $\tau_m$, is returned as the system's ultimate response ($\mathbf{r}$). By enforcing such a structured format and history-aware reasoning mechanism, SoccerAgent ensures robust, interpretable, and accurate responses to different multimodal soccer questions.

# 5 Experiments

This section starts from the description of experimental settings in Sec. 5.1, and presents extensive quantitative comparisons between SoccerAgent with state-of-the-art MLLMs in Sec. 5.2 and 5.3, followed by qualitative analysis in Sec. 5.4.

## 5.1 Experimental Settings

*5.1.1 Baselines.* We compare **SoccerAgent** against several representative MLLMs on **SoccerBench**, including both commercial APIs (Claude 3.7 Sonnet [1], Gemini 2.0 Flash [14], and GPT-4o [39]) and open-source models (DeepSeek-v3 [32], DeepSeek-R1 [17], Qwen2.5-VL [3], VideoLLaMA3 [65], LLaVA-Video [66], *etc*).

*5.1.2 Evaluation Metrics.* In all experiments, we compare the performance on multi-choice QA pairs in SoccerBench, and use the answer accuracy as the evaluation metric. We report both task-specific accuracy and category-specific (TextQA, ImageQA, and VideoQA) accuracy, to comprehensively reflect model performance.

**Table 2: Quantitative Comparisons on SoccerBench. Here, * indicates the use of a Commercial API (GPT-4o [39]) as a tool in the recommended tool chain for the corresponding task. "Open" denotes generating open-ended answers that are then converted to multi-choice options, while "MCQ" refers to inserting candidate options as context.**

| Model | TextQA | | ImageQA | | | | VideoQA | | | | | | | Overall | | |
|---|---|---|---|---|---|---|---|---|---|---|---|---|---|---|---|---|
| | Q1 | Q2 | Q3 | Q4 | Q5 | Q6 | Q7 | Q8 | Q9 | Q10 | Q11 | Q12 | Q13 | Text | Image | Video |
| *Commercial APIs* | | | | | | | | | | | | | | | | |
| Claude 3.7 Sonnet [1] | 58.1 | 58.2 | 51.3 | 32.0 | 63.3 | 63.9 | 39.8 | 26.8 | 48.3 | 49.3 | 38.6 | 43.9 | 45.5 | 58.1 | 47.1 | 43.4 |
| Gemini 2.0 Flash [14] | 61.9 | 52.2 | 63.2 | 41.0 | 88.5 | 67.3 | 59.0 | 46.0 | 56.1 | 62.7 | 42.8 | **52.4** | 55.0 | 57.6 | 56.5 | 54.0 |
| GPT-4o [39] | 64.0 | 58.5 | **76.7** | 46.0 | **89.6** | 70.6 | 61.3 | 40.0 | 66.4 | 70.0 | 43.7 | 49.9 | 59.7 | 61.6 | 62.3 | 57.5 |
| *Open-Source Models* | | | | | | | | | | | | | | | | |
| DeepSeek-v3 [32] | 56.0 | 49.5 | - | - | - | - | - | - | - | - | - | - | - | 53.1 | - | - |
| DeepSeek-R1 [17] | 68.3 | 51.1 | - | - | - | - | - | - | - | - | - | - | - | 60.6 | - | - |
| Qwen2.5-VL (7B) [3] | 35.6 | 53.5 | 58.5 | 35.8 | 82.0 | 66.0 | 56.8 | 31.6 | 52.2 | 51.6 | 35.0 | 46.9 | 50.7 | 43.6 | 52.4 | 46.8 |
| Qwen2.5-VL (72B) [3] | 49.4 | 37.7 | 66.5 | 45.9 | 87.0 | 67.5 | **67.5** | 19.5 | 58.8 | 58.5 | 51.0 | 49.0 | 58.7 | 44.2 | 59.3 | 53.2 |
| LLaVA-onevision (7B) [27] | 37.4 | 42.5 | 47.6 | 32.3 | 84.5 | 62.8 | 38.2 | 23.0 | 24.5 | 26.8 | 35.5 | 29.1 | 49.3 | 39.6 | 48.1 | 30.3 |
| VideoLLaMA3 (7B) [65] | - | - | 54.3 | 41.9 | 78.6 | 66.3 | 49.5 | 23.3 | 39.6 | 43.6 | 35.0 | 46.3 | 43.0 | - | 50.4 | 40.4 |
| LLaVA-Video (7B) [66] | - | - | 59.3 | 39.6 | 38.0 | 61.0 | 50.9 | 26.3 | 41.2 | 49.8 | 41.8 | 48.4 | 59.3 | - | 54.1 | 45.0 |
| VideoChat-Flash-Qwen2 (7B) [31] | - | - | - | - | - | - | 51.8 | 21.9 | 40.5 | 48.7 | **54.8** | 42.2 | 48.3 | - | - | 45.0 |
| **SoccerAgent (MCQ)** | **95.9** | 71.4 | 73.4* | **69.2** | 85.7 | 75.8 | 51.1 | 35.7 | 85.0 | **72.9** | 49.0 | 46.0 | 55.5 | **85.0** | **73.3** | 60.9 |
| **SoccerAgent (Open)** | 91.4 | 71.4 | 73.8* | 65.3 | 85.0 | 73.6 | 51.1 | 30.4 | 82.6 | 69.5 | 49.0 | 45.8 | 56.7 | 82.5 | 70.9 | 59.3 |
| **SoccerAgent-GPT4o (Open)** | 92.1 | **73.6** | 73.8 | 68.6 | 80.0 | **78.4** | 58.8 | **55.0** | **85.5** | 71.0 | 50.0 | 51.3 | **60.0** | 83.9 | **73.3** | **64.3** |

*5.1.3 Implementation Details.* We implement three different variants of SoccerAgent: (i) SoccerAgent (MCQ) takes both questions and candidate options as input to select the answer; (ii) SoccerAgent (Open) takes only the question as input, generates an open-ended answer, and then maps it to the provided multi-choice options; (iii) SoccerAgent-GPT4o (Open) replaces the open-source VLM tool (Qwen2.5VL-7B [3]) with GPT4o [39] to reflect the evolutionary potential of SoccerAgent. Moreover, all baseline methods follow the evaluation protocol aligned with SoccerAgent (MCQ).

## 5.2 Quantitative Results

According to the results presented in Table 2, we have the following observations of our SoccerBench: (i) the benchmark effectively differentiates the soccer understanding capabilities of existing MLLMs, with accuracy ranges spanning TextQA (39.6–61.6%), ImageQA (47.1–62.3%), and VideoQA (30.3–57.5%). This variation reflects the diverse and challenging nature of SoccerBench, as well as the varying levels of soccer-specific knowledge among existing models; (ii) distinct models excel in specific QA tasks, *e.g.*, GPT-4o achieves significantly higher performance in Q3-*Camera Status Classification*, Q5-*Jersey Number Recognition* and Q13-*Multi-view Foul Recognition*, while Gemini 2.0 Flash substantially outperforms in Q8-*Replay Grounding* and Q12-*Jersey Color Relevant QA*, highlighting their specialization in soccer understanding tasks; and (iii) most models perform well on tasks requiring less domain knowledge, *e.g.*, Q5-*Jersey Number Recognition* and Q7-*Camera Status Switching*, but still struggle with other tasks demanding in-depth soccer-specific knowledge. This indicates that current models are still not capable of fully handling comprehensive soccer understanding tasks. More results are provided in the **Appendix**.

In contrast to the above baselines, all three variants of SoccerAgent consistently outperform with the following characteristics: (i) superior performance on questions requiring soccer-specific knowledge, *e.g.*, Q1/4-*Background Knowledge Text/Image QA* and Q9-*Action Classification*; and (ii) leading results across TextQA, ImageQA, and VideoQA. Among them, SoccerAgent (MCQ) adopts the same evaluation protocol as all the baselines, ensuring a fair comparison. SoccerAgent (Open) first generates open-ended responses and subsequently selects final answers from multiple choices, as detailed in Sec. 5.1.3, crucially without access to the candidate options during reasoning and tool execution. Despite this constraint, it still achieves competitive performance comparable to SoccerAgent (MCQ), demonstrating robust open-ended QA capabilities of SoccerAgent. Moreover, SoccerAgent-GPT4o (Open), equipped with a more powerful VLM tool, performs even better, reflecting the scalability and evolutionary potential of SoccerAgent.

## 5.3 Ablation Studies

To systematically evaluate the intrinsic soccer understanding abilities, we conduct ablation studies on several variants of SoccerAgent (Open). Concretely, we consider: (i) whether to provide the planning agent ($\mathcal{A}_{plan}$) with task descriptions, including taxonomic definitions of all 13 question types and recommended tool chains; and (ii) whether to supply the execution agent ($\mathcal{A}_{exec}$) with 20 fully annotated execution examples demonstrating the optimal tool execution process. These quantitatively assess the autonomous soccer reasoning capacities of both components in SoccerAgent.

As depicted in Table 3, variations in task descriptions and execution examples have minimal impact on overall accuracy, indicating stable performance in both problem decomposition and tool execution. However, several noteworthy observations emerge: (i) The incorporation of text descriptions leads to accuracy improvements across most question types, highlighting the value of clear task explanations, especially for the planning agent ($\mathcal{A}_{plan}$); and (ii) While the extra execution examples yield a notable performance gain (+3.6%) in TextQA tasks, they slightly decrease accuracy for ImageQA (-0.4%) and VideoQA (-1.1%). This indicates that the execution module ($\mathcal{A}_{exec}$) is inherently adept at processing visual information effectively, whereas additional few-shot examples may introduce counterproductive effects. Thus, we choose to provide detailed task descriptions for SoccerAgent without introducing extra execution examples, as the default configuration.

**Table 3: Ablations on SoccerAgent (Open). Here, gray background indicates the default configuration of SoccerAgent, while TD and EX denote task descriptions and execution examples, respectively.**

| Strategy | | TextQA | | ImageQA | | | | VideoQA | | | | | | | Overall | | |
|---|---|---|---|---|---|---|---|---|---|---|---|---|---|---|---|---|---|
| TD | EX | Q1 | Q2 | Q3 | Q4 | Q5 | Q6 | Q7 | Q8 | Q9 | Q10 | Q11 | Q12 | Q13 | Text | Image | Video |
| ✘ | ✘ | 91.0 | 69.7 | 73.2 | 39.8 | 84.9 | 71.1 | 50.9 | 28.6 | **83.6** | 68.9 | 30.7 | 43.3 | **57.0** | 81.5 | 58.5 | 55.7 |
| ✔ | ✘ | 91.4 | 71.4 | **73.8** | 65.3 | **85.0** | **73.6** | **51.1** | **30.4** | 82.6 | **69.5** | **49.0** | **45.8** | 56.7 | 82.5 | **70.9** | **59.3** |
| ✔ | ✔ | **92.9** | **77.7** | 67.5 | **67.8** | 85.0 | 72.0 | 47.1 | 27.6 | 82.6 | 68.3 | 48.6 | 44.7 | 56.4 | **86.1** | 70.5 | 58.2 |

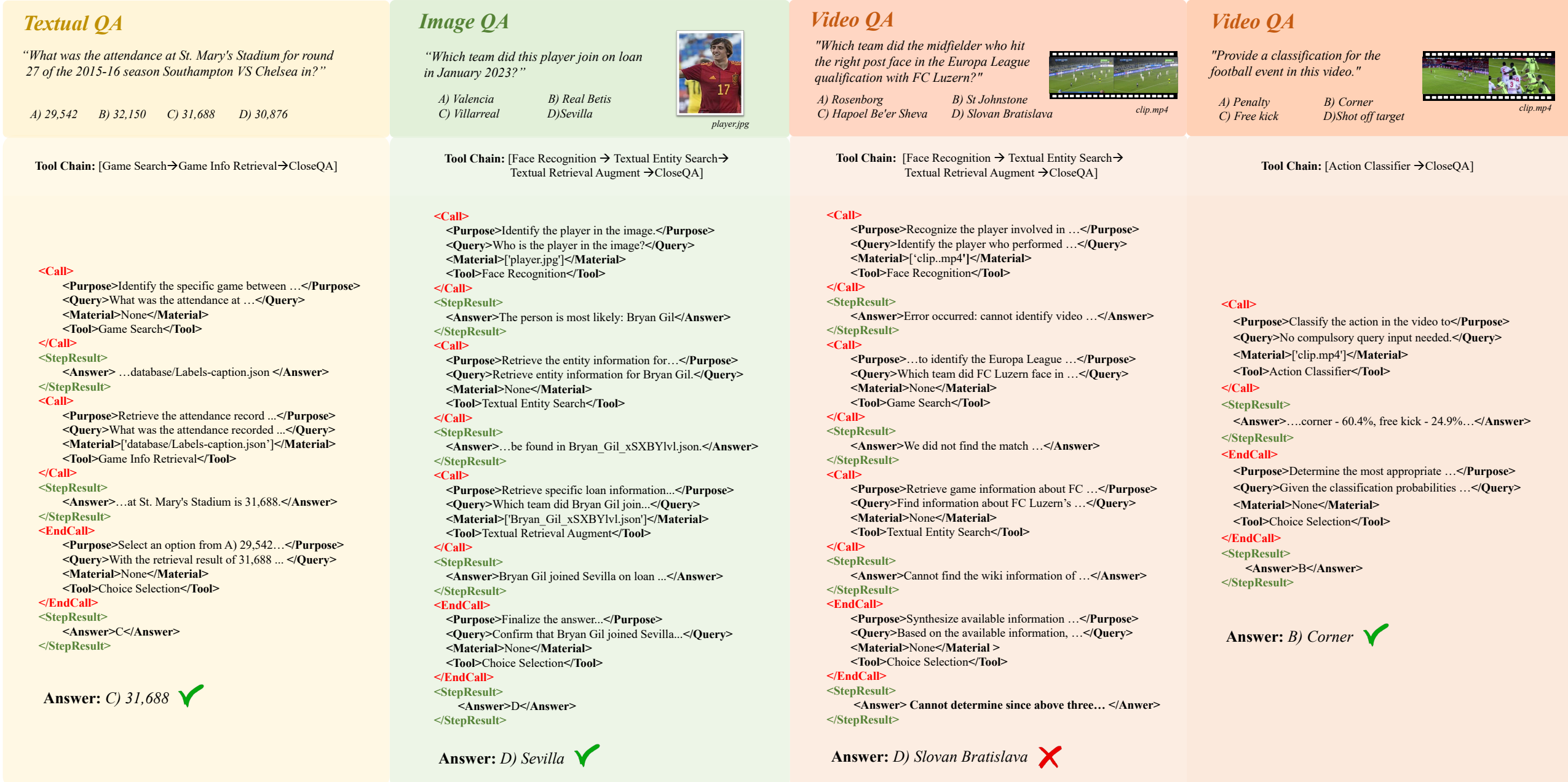

**Figure 4: Qualitative Results. Here, we demonstrate several representative examples showing the entire process of tool planning and tool execution of different soccer understanding tasks.**

## 5.4 Qualitative Results

As shown in Figure 4, we present representative qualitative results across TextQA, ImageQA, and VideoQA to illustrate the complete operation process of SoccerAgent, highlighting its reasoning logic, multimodal processing, and tool execution. In most cases, the framework seamlessly integrates tools, effectively decomposing and addressing soccer-related questions. Occasionally, execution errors occur: for example, in the third case, the *Face Recognition* tool is mistakenly invoked for a video input due to suboptimal tool chain planning. However, by integrating the results and historical context into the next-step planning process, $\mathcal{A}_{exec}$ autonomously adjusts its strategy, adopting the *Game Search* tool to retrieve the mentioned game. This showcases SoccerAgent's error-correction capability, advanced tool functionalities comprehension, and domain-specific expertise in soccer understanding. More experimental results and detailed analysis are provided in the **Appendix**.

## 6 Conclusion

This paper presents a comprehensive framework for holistic soccer understanding. Concretely, we introduce **SoccerWiki**, the first large-scale, soccer-specific multimodal knowledge base, integrating rich domain knowledge. Building on SoccerWiki and existing data sources, we present **SoccerBench**, the most comprehensive soccer understanding benchmark to date, featuring around 10K standardized QA pairs across 13 distinct tasks. To tackle this challenging and knowledge-intensive task, we establish **SoccerAgent**, a novel multi-agent system that achieves robust performance through collaborative reasoning and domain expertise. Extensive evaluations have demonstrated the superiority of our framework, establishing a new foundation for knowledge-driven sports analytics.

## Acknowledgments

Weidi would like to acknowledge the funding from Scientific Research Innovation Capability Support Project for Young Faculty (ZY-GXQNJSKYCXNLZCXM-I22).

# Multi-Agent System for Comprehensive Soccer Understanding

## Appendix

## A Comparison with other datasets

Our SoccerBench integrates a diverse collection of existing and newly curated datasets, establishing a unified framework for evaluating soccer understanding across multiple tasks. As illustrated in Table 4, we systematically compare the task coverage of current soccer-related benchmarks, highlighting both the breadth of our approach and gaps in prior work. This comprehensive benchmark enables cross-task evaluation while addressing the need for standardized assessment in soccer AI research.

**Table 4: Comparison of the task coverage of our SoccerBench against other soccer-related datasets.**

| Dataset | BackGround | Match Information | Camera Status | Jersey Number | Jersey Color | Replay Grounding | Action Classification | Commentary | Foul Recognition |
|---|---|---|---|---|---|---|---|---|---|
| SoccerNet-v1 [13] | ✘ | ✔ | ✘ | ✘ | ✘ | ✘ | ✘ | ✘ | ✘ |
| SoccerNet-v2 [7] | ✘ | ✔ | ✔ | ✘ | ✘ | ✔ | ✔ | ✘ | ✘ |
| SoccerNet-Caption [38] | ✘ | ✔ | ✘ | ✘ | ✘ | ✘ | ✘ | ✔ | ✘ |
| SoccerNet-JN [5] | ✘ | ✘ | ✔ | ✘ | ✘ | ✘ | ✘ | ✘ | ✘ |
| GOAL [40] | ✘ | ✔ | ✘ | ✘ | ✘ | ✘ | ✘ | ✔ | ✘ |
| SoccerNet-XFoul[22] | ✘ | ✘ | ✘ | ✘ | ✘ | ✘ | ✘ | ✘ | ✔ |
| SoccerReplay-1988[43] | ✘ | ✔ | ✘ | ✘ | ✘ | ✘ | ✔ | ✔ | ✘ |
| **SoccerBench(Ours)** | ✔ | ✔ | ✔ | ✔ | ✔ | ✔ | ✔ | ✔ | ✔ |

Table 5 presents a comparison of SoccerBench with other sports QA datasets. Our benchmark includes three types of tasks: text, image, and video QA, providing broader modality coverage than prior datasets.

**Table 5: A comparison of size between SoccerBench and other sports QA datasets**

| Dataset | Text | Image | Video |
|---|---|---|---|
| BIG-bench [48] on sports | ✔ | ✘ | ✘ |
| QASports [24] | ✔ | ✘ | ✘ |
| LiveQA [34] | ✔ | ✘ | ✘ |
| SportQA [58] | ✔ | ✘ | ✘ |
| SoccerNet-XFoul [22] | ✘ | ✘ | ✔ |
| Sports-QA [29] | ✘ | ✘ | ✔ |
| SPORTU [59] | ✔ | ✘ | ✔ |
| **SoccerBench(Ours)** | ✔ | ✔ | ✔ |

## B Additional Experiment Results

### B.1 More Results on the SoccerBench

We also evaluate Gemini 2.5 Pro Exp. [15] on SoccerBench, with the results presented in Table 6. Due to budget limitations, we test all TextQA and ImageQA tasks but restrict VideoQA evaluation to a subset of 100 QA pairs per task. The findings demonstrate that Gemini 2.5 Pro Exp. achieves strong performance across multiple tasks in SoccerBench, indicating its robust capability in soccer-related understanding.

**Table 6: Quantitative Results for Gemini 2.5 Pro on SoccerBench. Considering the cost, we randomly select 100 QAs as a subset for evaluation of the corresponding task.**

| Model | TextQA | | ImageQA | | | | VideoQA | | | | | | | Overall | | |
|---|---|---|---|---|---|---|---|---|---|---|---|---|---|---|---|---|
| | Q1 | Q2 | Q3 | Q4 | Q5 | Q6 | Q7 | Q8 | Q9 | Q10 | Q11 | Q12 | Q13 | Text | Image | Video |
| Gemini 2.5 Pro [15] | 77.5 | 71.0 | 70.9 | 64.0 | 89.5 | 83.8 | 68.0* | 94.0* | 69.0* | 71.0* | 61.0* | 48.0* | 58.0* | 74.6 | 73.0 | 66.2* |

### B.2 Cost Analysis

SoccerAgent adopts the official DeepSeek-v3 (671B) [32] API for agent cores, combined with locally deployed tools (UniSoccer, Qwen2.5VL-7B, etc.), requiring 40GB GPU memory. As for the average cost of it on SoccerBench: on an H800 GPU, each inference costs within $0.1 (API fee) for 20s (ranging from 5-30s), showing SoccerAgent's excellent deployability and scalability.

## B.3 Failure Analysis

SoccerAgent's failures primarily stem from tool planning and execution. Here, we track the performance of some intermediate steps to analyze system bottlenecks. (i) Planning failures mainly occur in complex tasks requiring multiple steps, such as Q4 and Q11 in Table 3, which is effectively mitigated with the utilization of TD (Task Descriptions) as it can significantly reduce planning failure rates to below 5%. (ii) Most of the errors originate from inherent limitations of tools. For example, Match Search tool accounts for about 90% of failures of Q2 and Action Classification tool explains nearly all failures in Q9. Moreover, Entity Recognition's 63.4% player ID accuracy directly determines the 65.3% success rate for Q4.

# C Further Implementation Details of SoccerBench

## C.1 Details in Open-ended QA Construction

As mentioned before, we derive specialized methods to generate open-ended soccer questions with three methods: (i) LLM Generation with DeepSeek-V3 [32]; (ii) Template applications; and (iii) Both LLM and Template. For each task in SoccerBench, we decide the question generation method considering the form and characteristics of SoccerWiki or other source datasets. Notably, for both textual commentary and event labels, we utilize only the test sets from the respective datasets to avoid potential data leakage to the trained models used later. The detailed introduction is listed as follows:

*C.1.1 Background Knowledge Text QA.* With the textual information from wiki pages of all the players, teams, referees, and venues, we use DeepSeek-V3 [32] to generate questions from the entire page. The generated question should be equipped with its answer and the reference position in the wiki page to ensure the rationality of the generated questions.

*C.1.2 Match Situation QA.* Similar to *Background Knowledge Text QA*, we use the same method to generate questions from the existing soccer datasets, MatchTime [43] and SoccerReplay-1988 [44]. With `.json` game files as input, LLM would generate questions about the game events and relevant information.

*C.1.3 Camera Status Classification.* We use the 13 categories of camera status annotations in SoccerNet-v2 [7] to capture the corresponding images in the match video. Then we use templates such as "What is the camera position in this picture?" to directly ask the camera status to construct QA.

*C.1.4 Background Knowledge Image QA.* Based on *Background Knowledge Text QA*, we first find the picture corresponding to the player's name in the question in SoccerWiki, and then use the pronoun "this player" to replace the name in the question to complete this construction.

*C.1.5 Jersey Number Recognition.* We utilize the player images and corresponding jersey number ground truth annotations in SoccerNet-JN [5], using templates like "What is the number on the player's jersey in this image?" to directly ask the jersey number in the image.

*C.1.6 Score and Time Relevant QA.* Basically, this task provides a screenshot image of a soccer broadcast with time and scoreboard shown inside. With this digital information, 2 different types of questions could be generated: (i) *Time and Score Recognition*: For any second of a soccer game, the game time and scores could be reached from soccer commentary datasets, MatchTime [43] and SoccerReplay-1988 [44]. We derive a template to generate such questions and make screenshots from according games with corresponding time stamps. (ii)*Retrieval Required Questions:* With game time and game events available, we introduce the basic game information until this game time to DeepSeek-V3 [32] and let it generate the questions corresponding to game time and game information.

*C.1.7 Camera Status Switching.* Similar to *Camera Status Classification*, we use the camera status annotations in SoccerNet-v2 [7] and cut a video clip before and after the camera switch. Directly ask the switch type by templates like "What kind of camera transitions are used in the video?" to construct QA.

*C.1.8 Replay Grounding.* Using the replay annotations from SoccerNet-v2 [7], we first extract replay video clips based on the provided timestamps. Next, we use the link annotations to extract the action video clips corresponding to each replay, thereby constructing a replay-action relationship pair. Then we use templates like "The first video clip is a replay. From the remaining clips, please choose which one is being replayed." to construct QA.

*C.1.9 Action Classification.* With plenty of event labels in MatchTime [43] and SoccerReplay-1988 [44], we set some templates to generate questions requiring the event label in the video clips.

*C.1.10 Commentary Generation.* As for *Commentary Generation*, the methodology is the same as *Action Classification*.

*C.1.11 Commentary Relevant QA.* We could capture the player name from non-anonymized commentaries in MatchTime [43] and SoccerReplay-1988 [44], then we send the player name and his generated questions in *Background Knowledge Text QA* to DeepSeek-V3 [32] to generate the question combining the commentaries about the player himself. Such questions could share the same answer as questions in *Background Knowledge Text QA*.

*C.1.12 Jersey Color Relevant QA.* With our manually labeled jersey colors of both teams for all the games in MatchTime [43] and SoccerReplay-1988 [44]. DeepSeek-V3 [32] helps generate questions combining the information of jersey color and commentaries.

*C.1.13 Multi-view Foul Recognition.* As SoccerNet-XFouls [21] has labeled different dimensions of soccer fouls with multi-view foul video clips. We directly use its template and options to generate questions on specific dimensions (*e.g.*, handball, foul class, *etc.*)

## C.2 Prompt to Convert Open-ended QA to Multi-choice QA

Convert this soccer-related open QA pair into a multiple-choice question with four options:
Original Question: {question}
Reference Answer: {answer}
Requirements:
1. Maintain the question's core meaning while adapting it for multiple-choice format
2. Generate four options where:
- The correct option MUST exactly match the reference answer
- Distractors should:
* Be thematically relevant and plausible
* Avoid obvious errors (e.g., incorrect club names)
* Include common misconceptions or easily confused elements
* Maintain consistent granularity (e.g., same era comparisons)
* Use parallel structures and similar lengths
* Avoid grammatical cues that reveal the answer
3. Randomize option order (correct answer not fixed to any position, you must randomize the order of the options)
4. Use this exact format:
Q: [Question text]
A: [Correct option letter]
O1: [Option 1]
O2: [Option 2]
O3: [Option 3]
O4: [Option 4]
Example 1:
Q: Which player holds the record for most goals in a single Champions League season?
A: O3
O1: Cristiano Ronaldo
O2: Lionel Messi
O3: Robert Lewandowski
O4: Karim Benzema
Example 2:
Q: What is the maximum number of substitutes allowed in a standard Premier League match?
A: O2
O1: 3
O2: 5
O3: 7
O4: 9
Example 3:
Q: Which country won the first FIFA World Cup in 1930?
A: O4
O1: Brazil
O2: Germany
O3: Italy
O4: Uruguay
Example 4:
Q: What is the minimum stadium capacity required for hosting a FIFA World Cup final?
A: O1
O1: 80,000
O2: 60,000
O3: 50,000
O4: 40,000

**Table 7: League distribution of the SoccerBench Q1 and Q2.**

| League | Q1#Player | Q1#Team | Q2#Game |
|---|---|---|---|
| Premier League | 207 | 20 | 268 |
| La Liga | 129 | 14 | 144 |
| Serie A | 168 | 18 | 159 |
| Ligue 1 | 85 | 12 | 149 |
| Champions League | 319 | 32 | 249 |
| Bundesliga | 107 | 16 | 131 |
| Non-Europe Leagues | 37 | - | - |

**Table 8: Seasonal distribution of the SoccerBench Q1 and Q2.**

| Season | Q1#Player | Q2#Game |
|---|---|---|
| 2014–2015 | 20 | 46 |
| 2015–2016 | 17 | 61 |
| 2016–2017 | 74 | 103 |
| 2017–2018 | 63 | 71 |
| 2018–2019 | 138 | 132 |
| 2019–2020 | 138 | 107 |
| 2020–2021 | 110 | 131 |
| 2021–2022 | 166 | 131 |
| 2022–2023 | 215 | 173 |
| 2023–2024 | 102 | 45 |

## C.3 Details in Data Selection

As mentioned in Sec 3.3, we selected around 10K representative samples from the initially synthesized 100K QA pairs to form **SoccerBench**. Our selection process employs the following criteria: we first randomly shuffle all data and verify question-answer correspondence according to metadata; then we ensure uniform sampling of entities (e.g., players, teams, leagues, and seasons) to ensure a balanced data distribution, as detailed in Sec C.4. Subsequently, we manually filter out potentially low-quality samples (e.g., questions in ImageQA and VideoQA that can be answered without visual context).

## C.4 Data Distribution

To ensure that **SoccerBench** serves as a diverse and balanced benchmark, we strived to maintain a well-distributed dataset during the data selection process. For example, our **TextQA Q1** (1,500 questions) spans 1,121 players (93.3% appearing only once, with just 4 players being queried three times), 100 teams, as well as some referees and stadiums. Additionally, in **TextQA Q2**, we also balance the distribution of games across leagues and seasons, as shown in Table 7 and Table 8.

While our dataset indeed primarily focuses on European leagues (2014-2024) due to limited available structured data, we have also included (non-European) player information from the last 3 World Cups (2014/2018/2022) to ensure broader diversity.

# D Further Implementation Details of SoccerAgent

## D.1 Task Description Prompt (with recommended chains)

**Task1:** **Background knowledge text QA** is a task that asks questions about the basic information of a specific players, referee, team and venues. Ask about those questions could be answer from their WikiPage
Recommended chain: Textual Entity Search -> Textual Retrieval Augment -> LLM
**Task2:** **Match Situation QA** is a task that asks questions about the basic information of a specific match, the players and coaches of both teams, and important events of the match (goals, assists, red and yellow cards, etc.) The game range only covers 2014-2024's six european major leagues (Premier, Bundesliga, Serie-a, Ligue-1, Laliga and European Champions League)

Recommended chain: Game Search -> Game Info Retrieval -> Match History Retrieval -> LLM
**Task3:** **Match Events and Statistical QA** is a task to ask questions about the history events of the match. For example, 'How many corners has xxx team get in the first half.'. The game range only covers 2014-2024's six european major leagues (Premier, Bundesliga, Serie-a, Ligue-1, Laliga and European Champions League)
Recommended chain: Game Search -> Game Info Retrieval -> Match History Retrieval -> LLM
**Task4:** **Camera Status Classification** is a task that determines the state of the camera position in the picture at a certain moment in the game.
Recommended chain: Camera Detection -> LLM
**Task5:** **Background knowledge Image QA** is a task that asks questions with one or more images about the basic information of a specific players, referee, team and venues.
Recommended chain: Entity Recognition -> Textual Entity Search -> Textual Retrieval Augment -> LLM
**Task6:** **Jersey Number Recognition** is a task to identify the jersey numbers of a players in images.
Recommended chain: Number Recognition -> LLM
**Task7:** **Score and Time Relevant QA** is a task asking about questions that starts from scores or gametime, which means you need to recognize the time or score from the given materials of soccer broadcast. Sometimes you need to know game information and sometimes you only need to recognize and then answer the question.
Recommended chain: Score and Time Recognition -> LLM
**Task8:** **Camera Status Switching** is a task to judge the state of the camera position switching in the video clip.
Recommended chain: Shot Change -> Camera Detection (twice) -> LLM
**Task9:** **Replay Grounding** is a task to identify which video clip is being replayed from a set of clips, with the first clip serving as the replay.
Recommended chain: Commentary Generation (five times) -> LLM
**Task10:** **Action Classification** is a task to classify the actions of the events on soccer game in the video clip.
Recommended chain: Action Classifier -> LLM
**Task11:** **Commentary Generation** is a task to generate commentary for the events in the video clip.
Recommended chain: Commentary Generation -> LLM
**Task12:** **Commentary Relevant QA** is a task to ask questions about background information of certain player with the question having commentary descriptions.
Recommended chain: Vision Language Model -> LLM
**Task13:** **Jersy Color Relevant QA** is a task to ask questions about soccer stuffs like players, matches. All these questions are with elements of jersey colors.
Recommended chain: Vision Language Model -> LLM
**Task14:** **Multi-view Foul Recognition** is a task to recognize the fouls in the video clip from multiple views.
Recommended chain: Foul Recognition -> LLM

## D.2 Tool Description

=== Tool Description for TOOL1 ===
Name: Choice Selection
Ability: Given an open-ended answer to a question, the tool identifies the most appropriate answer choice from a set of closed-ended (multiple-choice) options. It analyzes the open answer and matches it to the correct option.
Query Input: A query containing the question and its according options (in forms of 'o1', 'o2', ......), together with an answer which is generated already as an openQA answer
material Input: No material input is acceptable, or some relevant file could also be input
Output: Considering the question and the openQA answer, the option will be generated finally.
Remark: This tool is only used for those CloseQA settings, especially when the openQA answer is already generated.
=== Tool Description for TOOL2 ===
Name: LLM
Ability: Given a prompt, the tool can perform a variety of natural language tasks such as text generation, question answering, summarization, sentiment analysis, translation and more, leveraging the power of a large-scale pre-trained language model. You can use it as a tool of solving textual problems.
Query Input: A prompt that has clear requirement, and better to have define the output form material Input: No file material needed.
Output: The response according to the prompt.
Remark: Just understand this tool as a powerful language model, which can be used to solve various textual problems.

=== Tool Description for TOOL3 ===
Name: Action Classifier
Ability: Given a video clip, the tool classifies the actions of the soccer event to one of the 24 predefined types.
Query Input: No compulsory query input needed, the question setting could be provided as the query input.
material Input: A list with first element is the file path to a video.
Output: Output one or more categories that are most likely to be the classified event type of the video in material input.
Remark: This step can generate the most probable types of actions in the video. With normally over 80 percents of top-1 accuracy.
=== Tool Description for TOOL4 ===
Name: Commentary Generation
Ability: Given a video clip and game context, the tool generates commentary text based on the events in the video clip. Such commentary is anonymized with '[PLAYER]', '[TEAM]', '[REFEREE]', '[COACH]' for according entities
Query Input: No compulsory query input needed, the question setting could be provided as the query input.
material Input: A list with first element is the file path to a video.
Output: Output a commentary that can describe the soccer event happened in the video of material input.
Remark: CIDEr Score to ground truth around 0.2-0.5.
=== Tool Description for TOOL5 ===
Name: Foul Recognition
Ability: Given multi-view foul video clips, the tool recognizes the fouls in the soccer match and classifies them into different categories severity from 1 to 5 and select according foul type, etc.
Query Input: The question setting could be provided as the query input.
material Input: A list of videos that are all multi-view foul video clips of same foul.
Output: The foul type and severity will be provided.
Remark: severity from integer 1 to 5, and the type of the foul.
=== Tool Description for TOOL6 ===
Name: Game Search
Ability: Given some information of a match, the tool retrieve which game it is from soccer match database. The games are from 6 European major legues (England Premier, Germany Bundesliga, Italy Serie-a, Spain Laliga, France Ligue-1 and European Champions League) during 2017-2024.
Query Input: Just the original question as query input here, containing some game information.
material Input: No compulsory file path needed.
Output: The JSON file paht of the retrieved game. Or if no matching file, will response accordingly.
Remark: This tool must be done at first to get the game context if you want to know the game information (e.g. who is the referee, how many attendance, how many corner kick in total, .etc), and then other tools can be used with such JSON file.
=== Tool Description for TOOL7 ===
Name: Game Info Retrieval
Ability: Knowing the game context, the tool retrieves the game information from the soccer match database. Such information specially refers to those information that could be know before the match kick off moment (e.g. the referee, coach, the attendance, the foramtion) and final results like final scores.
Query Input: Query input could be the original question, or the well defined question that can help retrieve the question.
material Input: A list with first file (always only one file) is JSON file path provided by Game Search.
Output: The answer to query input considering the game contents from the JSON game file.
Remark: This tool is always used after Game Search, and the game information is always provided in the JSON file. This tool is always done sequentially with 'Match History Retrieval' with same query and material input so that the total match info would be retrieved.
=== Tool Description for TOOL8 ===
Name: Match History Retrieval
Ability: Knowing the game context, the tool retrieves the match history information from the soccer match database. Such match history is always the textual live stream of whole gamme in a JSON file from Game Search tool.
Query Input: Query input could be the original question, or the well defined question that can help retrieve the question.
material Input: A list with first file (always only one file) is JSON file path provided by Game Search.
Output: The answer to query input considering the game contents from the JSON game file.
Remark: This tool is always used for retrieve some information of the game process itself or some statistics of the game. This tool is always done sequentially with 'Game Info Retrieval' with same query and material input so that the total match info would be retrieved.
=== Tool Description for TOOL9 ===

Name: Textual Retrieval Augment
Ability: Given a text query, the tool retrieves the relevant information from given soccer information or database page.
Query Input: Prompt query could be the original question, or the well defined question that can help retrieve the question.
material Input: A list with first file (always only one file) is JSON file path provided by Game Search.
Output: The answer to query input considering the game contents from the JSON game file.
Remark: This tool is always used for retrieve information except above two tools. It's always be used for background information of players, teams, coaches, referees, venues, etc. You can understand it as a retrieval tool of a huge soccer background database.
=== Tool Description for TOOL10 ===
Name: Textual Entity Search
Ability: Given question about and entity(player, team, etc.), the tool retrieves the requiring entity of the question, and return its according WikiPage. The entity database contains the history and background knowledge for all the players, teams, venues, coaches and referees from games are from 2022 World Cup and 6 European major legues (England Premier, Germany Bundesliga, Italy Serie-a, Spain Laliga, France Ligue-1 and European Champions League) during 2017-2024.
Query Input: Prompt query could be the original question.
material Input: No compulsory material input needed.
Output: A list with first file (always only one file) is JSON file path containing the according entity's information.
Remark: Here is an important part that if the retrieval requirement is out of the game range in 'Game Search' database, you need to find that knowledge here to identify entity first, then retrieve the background knowledge.
=== Tool Description for TOOL11 ===
Name: Number Recognition
Ability: Given one or more images, detect and recognize the jersey number of the player present in the images.
Query Input: No compulsory query input needed, the question setting could be provided as the query input.
material Input: A list containing paths of a player's images.
Output: Jersey number of the player in the images. If no jersey number is detected, the result is -1.
Remark: If you want to know the jersey number of the player in the picture, please use the Number Recognition tool.
=== Tool Description for TOOL12 ===
Name: Camera Detection
Ability: Given one image or one video, the tool identifies and classifies the type of camera positions within the image or video among 13 camera types (e.g., Main Camera Center, Close-up player or field referee, Close-up Behind the Goal, etc.).
Query Input: No compulsory query input needed, the question setting could be provided as the query input.
material Input: A list with first element is the path to a match image or video.
Output: The camera position in the image. There are 13 possible results.
Remark: If you want to know the camera position of a specific frame or a video clip in the game, please use the Camera Detection tool.
=== Tool Description for TOOL13 ===
Name: Replay Grounding
Ability: Given more than one video clips, the tool assumes that the first one is a replay video and determines the clip being replayed from the next four clips.
Query Input: No compulsory query input needed, the question setting could be provided as the query input.
material Input: A list with five video paths. The first element is the path to a replay clip. The remaining four are possibly being replayed video clip paths.
Output: The path of video clip being replayed.
Remark: If you want to find the corresponding replay video clip from a set of video clips, please use the Replay Grounding tool.
=== Tool Description for TOOL14 ===
Name: Entity Recognition
Ability: Given one or more images, the tool identifies and recognizes the name of the player present in the images through face matching.
Query Input: No compulsory query input needed, the question setting could be provided as the query input.
material Input: A list containing paths of a player's images.
Output: The name of the player in the image. If no face is detected or no matching player is found, returns 'None'.
Remark: If you have some pictures of players and want to know who is in the picture, you can use Entity Recognition tool.
=== Tool Description for TOOL15 ===
Name: Jersey Color Relevent VQA
Ability: Given an image/video and a text query about soccer jersey, the tool generates answers or descriptions related the jersey color relevant QA answers. You can obtain any information about jersey (color) from this tool to help you understand soccer.

Query Input: A text prompt describing the information you want to know about the jersey color in this image/video.
material Input: A list containing the path of a single image, a sequence of images, or a video.
Output: The response exactly answer the jersey relevant questions.
Remark: This tool is required when you need any information about soccer and you don't have that, during your reasoning and QA process.
=== Tool Description for TOOL16 ===
Name: Segment
Ability: Given an image and a text description of the object you want to segment, the tool will get the bounding box coordinates of the object and the corresponding confidence score.
Query Input: A text description of the object you want to segment. The description should be as concise as possible and clear in direction.
material Input: A list with first element is the path to a image you want to segment.
Output: The bounding box coordinates of the object you want to segment and the corresponding confidence score.
Remark: If you want to segment an object (such as a player) in a photo to get a partial image, you can use Segment tool.
=== Tool Description for TOOL17 ===
Name: Score and Time Recognition
Ability: Given a video clip of match, the tool recognizes the score and time of the game from the soccer broadcast video clip.
Query Input: Give a query input about what you exactly want to know about score or game time, the question setting could be provided as the query input.
material Input: A list with single elements of image or a video clip from soccer game broadcast.
Output: Output the score and the game time shown in the file screenshot. If more than 1 picture was provided, return these information one by one.
Remark: This tool is used to recognize the score and time of the game from the soccer broadcast video clip, with image as input and text as output.
=== Tool Description for TOOL18 ===
Name: Frame Selection
Ability: Given a description query and a video of soccer game, the tool would select the frame that best match the prompt and save that frame as an image to certain path. Such image could be used for later steps.
Query Input: A prompt describing the frame that you want to obtain from the video.
material Input: A list with single element of file path to the video path.
Output: The file path of the saved image frame selected from the video according to the query prompt.
Remark: If the next step needs compulsory input of image but you only have video. This tool would be helpful.

### D.3 Task Decomposition Prompt

# Soccer Question Answering Assistant
## Task overview
You are a multi-modal agent that can answer questions about soccer knowledge.
For each question, you will receive:
- A question about soccer considering different aspects of soccer
- You might also receive one or more video clips or images as context
Your task involves three sequential parts:
1. Problem Decomposition (Part 1)
- Identify available information
- Break down the question into sequential steps
2. Sequential Tool Application (Part 2)
- Execute one tool at a time
- Record each tool's output
- Continue until sufficient information is gathered
3. Solution Synthesis (Part 3)
- Integrate all results
- Generate final answer
## Available Tools
For all the QA, you need to decompose them and Here are the tools that you can use to answer the questions:

```
{toolbox_descriptions}
## Common QA Tasks
Here are some common QA tasks that you might meet in the questions, for each types of questions, we provide the recommended tool chain for you to answer the questions:
{tasks}
To be noted, at this stage you only need to treat this question as open-ended QA task, you can use the common QA tasks as reference to decompose the question and identify the required tools.
## Response Format for Part 1
For each query, you should respond ONLY with:
Known Info: [list any categories explicitly mentioned in the query and material]
Tool Chain: [list required tools connected by ->]
## Examples
Query 1: "How does the viewpoint of the camera shift in the video?" Adittional Material: "video": $["clip.mp4"]$
Your response:
Known Info: [$VideoClip$]
Tool Chain: [*Shot Change* -> *Camera Detection* -> *LLM*]
Query 2: "What was the final score of the game 2015-02-21 - 18-00 Chelsea vs Burnley?"
Adittional Material: None
Your response:
Known Info: [$GameContext$]
Tool Chain: [*Game Search* -> *Game info Retrieval* -> *Match History Retrieval* -> *LLM*]
Query 3: "How many goals did the player who forced a corner score for Borussia Dortmund's senior team?"
Adittional Material: "video": $["clip.mp4"]$
Your response:
Known Info: [$VideoClip$, $GameContext$]
Tool Chain: [*Vision Language Model* -> *Entity Recognition* -> *Text Retrieval Augment* -> *LLM*]
## Important Rules
1. You should only use the tools provided in the toolbox to answer the questions and provide the exact tool names.
2. Use exact item category names with $$ to represent the information categories.
3. Use exact tool category names with ** as shown above to represent the tools.
4. Only respond with Part 1 analysis - Parts 2 & 3 will be addressed in subsequent interactions.
5. Connect tools using -> symbol
6. Try your best to decompose the question and identify the required tools, you can first reference the common QA tasks to get some ideas. If the template fits the question, you can directly use the recommended tool chain. If not, you can try to decompose the question and identify the required tools.
```

## D.4 Excution Prompt

```
As a multi-agent core in the Soccer Question Answering Assistant, you are required to execute the following tool chain to answer the question:
"{query}"
with the following additional material:
{material}
with the known info as:
{parse_input(response)[0]}
and you should execute the following tool chain to solve the question:
{parse_input(response)[1]}
As for the usage of the tools, you should follow the following references:
{toolbox}
For every tool above, we would input queries and materials into the tool for execution, the queries are in **text** form and the materials are in list with **file paths**. If no file path is suitable, you just write in 'None' You should determine the contents of materials and queries based on the context of the question, known info and tool descriptions.
For every steps of excution, you should return me with a clear statement of the goal of this step in the context of the overall analysis, the specific tool you are using, and the input variables you are using.
```

<Call>
<Purpose>Brief, clear statement of this step's goal in context of overall analysis</Purpose>
<Query>[Query/question here(string). IMPORTANT!!: Such query is highly relevant to the toolbox descriptions. you need to think carefully about your purpose this step and generate appropriate query.]</Query>
<Material>[Material list here(a string showing list form). Here as well, you need to think carefully considering the purpose and toolbox.]</Material>
<Tool>[Tool name here(string)]</Tool>
</Call>
If it is the last step of the execution, you should return me with the following format:
<EndCall>
<Purpose>Brief, clear statement of this step's goal in context of overall analysis</Purpose>
<Query>[Query/question here(string)]</Query>
<Material>[Material list with file paths here(a string showing list form)]</Material>
<Tool>[Tool name here(string)]</Tool>
</EndCall>
Every time you return me with the instruction as above, I will execute it and return you with the feedback of the execution in this format:
<StepResult>
<Answer>[The results of this time's execution here(string)]</Answer>
</StepResult>
For every time of generation, you should follow the following rules:
1. You should be clear about the tool name (must be chosen from toolbox), file path and query/question in the instruction. This part is important for me to understand the context of the execution. You cannot change any of the information in the instruction.
2. If I have given you the feedback of the execution, you should analyze what you should write in the next call based on the feedback considering the tool chain I gave you and the task descriptions and tool descriptions. You should not repeat the same instruction again.
3. If my prompt leaves you to generate the first call, you should directly return me with the call in the form from <> to </>. You should not add any other information in the instruction.
4. Otherwise, if in the prompt I have given you some <StepResult>, you should consider the total process of the execution and continue to return me exactly with the form from <> to </>. You should not add any other information in the instruction.
Once again, I repreat that the question is:
"{query}"
with the following additional material:
{material}
with the known info as:
{parse_input(response)[0]}
and you should execute the following tool chain to solve the question:
{parse_input(response)[1]}
The following is all our execution history, now you can start with your call of first step:

## D.5 Toolbox Prompt

**Game Search:**

You are a helpful assistant that extracts structured information from natural language text about football matches. I will give you a sentence about a football match, and you need to extract the following information: league, season, date, time, and two teams. The output must strictly follow the format below:
league: (england_epl, germany_bundesliga, europe_uefa-champions-league, italy_serie-a, france_league-1, spain_laliga, or unknown)
season: xxxx-xxxx
date: xxxx-xx-xx
year: xxxx
month: xx
day: xx
time: xx:xx (which means when this game kick-off, not the game timestamp of certain event)

score: x - x (if score is not determined, write 'unknown' for only in this attribute)
team1: yyy
team2: yyy
All above 'x' means a digit!! 'yyy' means a string.
To be noted, if you can determine only one team, please assign the team to team1 and leave team2 as 'unknown'. If any information is missing or uncertain, write 'unknown'. You have to use the exactly same name of teams as provided in the input text. Do not output any other words. For other attributes, if any information is missing or uncertain, write 'unknown'. As for date, you should record in the form of xxxx-xx-xx if you can get the clear date; Meanwhile, as for year, month, day, you need capture as more information point to this game as possible, including year, month, and day, and record them in numbers. Do not guess any information. For example if year is not said clearly, don't guess the year through season. Only use the information provided in the input text. Do not output any other words.

You are a helpful assistant that selects the most likely match from a list of candidates based on the given information. Now we need to retrieve a file path for the most probable match from the database from the question: "{question}".
Such question has been transformed to the original query information as:
{info}
Here are the candidate matches:
Candidate i:
- League: row['league']
- Season: row['season']
- Date: row['date']
- Year: row['year']
- Month: row['month']
- Day: row['day']
- Time: row['time']
- Score: row['score']
- Home Team: {row['home_team']}
- Away Team: {row['away_team']}
- file_path: {row['file_path']}

Based on the original query information and the candidate matches above, is there a match that is significantly more likely than the others?
Firstly, you should exclude those candidates in the following situation:
1. If **any of the team's name in original query information** is sure not to be in team names from candidates, such candidate cannot be returned anymore, you cannot let such candidate take place in your return answer.
2. For example, if the original query information contains "Chelsea" and "West Ham", but candidates contains "chelsea FC" and "Liverpool", since such candidate cannot be returned anymore since West Ham is not in candidate information.
3. For example, if the original query information contains "Chelsea" and "West Ham", but candidates contains "Chelsea FC" and "West Ham United", since such candidate is still possible to be returned since both team names are in candidate information.
4. For example, if the original query information contains only "Chelsea", but candidates contains "Bayern Munich" and "Real Madrid", since such candidate cannot be returned since Chelsea is not in candidate information.
After considering the above situation and exclude those candidate having team name unmatched, you should consider the following two situations:
1. If there are still **obviously** probable answer with all known information correct, please return the file path of that match EXACTLY in the following format: "The given information seems incomplete, but we found the most probable match in the database with this file path: [The file path of the **hugely most probable** match]. [Here give some recommendation to complete the information if possible, for example, provide the date or the score of the match, or which team is the home/away team etc. Use simple and clear words here.]"
2. If no match is significantly more likely among all the candidates, please return all candidate matches with information of league, season, date, time, score, home team, away team, venue and referee (without file path), and explain that the information provided is too vague. For this situation you only need to summarize with a little bit the games and give a brief reply with some short sentences.

**Entity Search:**

You are an intelligent assistant that can analyze questions related to football. Your task is to identify the type of entity mentioned in the question and extract the exact name of the entity. The entity types are: player, referee, team, venue. If the entity is a coach, classify it as a player. The name extracted should match exactly as it appears in the question.
Output the result strictly as a tuple in the format: (type, name). Do not include any additional explanations, notes, or formatting.
For example:
- Question: "How many goals did Lionel Messi score last season?"
Output: ("player", "Lionel Messi")
- Question: "Where is the Camp Nou stadium located?"
Output: ("venue", "Camp Nou")
- Question: "What was the decision made by referee Michael Oliver in the last match?"
Output: ("referee", "Michael Oliver")
- Question: "How did Manchester United perform in the last game?"
Output: ("team", "Manchester United")
However, if the entity type and entity name cannot be determined, please output as: ("unknown", "unknown")
For example:
- Question: "Explain the 4-4-2 formation." Output: ("unknown", "unknown")
- Question: "Who is the player in this image?" Output: ("player", "unknown")

**Match History Retrieval:**

Here is a question about soccer game:
"{query}"
The match history information has been found as following shows, you need to answer the question based on the information provided:
{match_history}
Please provide the answer based on the match history information. Please think it carefully and make sure your answer is evidence-based and accurate. Now answer the question in the following format:
[ANSWER]: [Your answer here] [EXPLANATION & REASONING]: [Your explanation here]
You should return exactly in this form without any other words.

**Game Info Retrieval:**

Here is a question about soccer game:
"{query}"
The match related information has been found as following shows, you need to answer the question based on the information provided:
{match_info}
Please provide the answer based on the match related information. Please think it carefully and make sure your answer is evidence-based and accurate. Now answer the question in the following format:
[ANSWER]: [Your answer here] [EXPLANATION & REASONING]: [Your explanation here]
You should return exactly in this form without any other words.

**Choice Selection:**

You are a football expert. You are provided with a question 'Q' and four options 'O1', 'O2', 'O3', and 'O4'.
Before I have used a helpful soccer multi-agent system to solve this process, I will tell you the total process of how agent deal with this problem.
Please answer the question with one option that best matches the question (replay with 'O1', 'O2', 'O3', or 'O4').
Do not include any other text or explanations!!!
This football question is "question". The four corresponding options are:
{options_str}
The processing through the multi-agent platform is as follows:

{openA_process}
Please provide your answer:

## D.6 Prompt of Soccer-specific Image Understanding Tools

**Camera detection:**

What is the camera position in this picture? The answer should be chosen from the following options: [Main camera center, Close-up player or field referee, Close-up side staff, Main camera left, Main behind the goal, Close-up behind the goal, Spider camera, Main camera right, Public, Goal line technology camera, Close-up corner, Inside the goal, Other].

**Jersey Number Recognition:**

Analyze this image and determine if the player is facing away from the camera. If the player is facing away, output the jersey number on their back. If the player is not facing away from the camera, output 'No'.

**Score/Time Recognition:**

What time is it in this soccer video? And what's the score?